**Кромер В.В.**
Новосибирск

# ИСТОРИЧЕСКАЯ ДИНАМИКА ЛЕКСИЧЕСКОЙ СИСТЕМЫ КАК ПРОЦЕСС СЛУЧАЙНОГО БЛУЖДАНИЯ

1. В настоящем докладе подводятся промежуточные результаты исследования математической модели исторической динамики лексической системы языка. Модель основывается на следующих положениях:

а) каждый синхронный срез лексической системы языка характеризуется некоторым полисемическим распределением лексики, а именно: в каждой полисемической зоне $m$ ($m$ – кол-во значений у всех слов зоны; $m = 1, 2, 3…$) присутствуют $n_m$ слов-лексем;

б) полисемическое распределение лексики задается ранее предложенной автором доклада моделью полисемического распределения лексики исходя из количества слов в лексической системе и суммарного количества значений [1];

в) лексическим системам разного объема свойственна инвариантная типологическая относительно полисемии характеристика языка [2, с. 29].

2. В данном докладе рассматривается стационарное состояние лексической системы. Стационарная лексическая система находится в состоянии динамического равновесия – приток новых слов в систему и удаление устаревших слов взаимно уравновешены. Наличие в лексической системе дискретных уровней – зон с равной степенью полисемии, определяемой количеством значений у слов – позволяет рассматривать жизнь слова в системе как случайные блуждания по дискретным уровням. Принципы случайного блуждания:

а) слово возникает на уровне 1, т.е. при возникновении попадает в зону моносемантичных слов (слов с одним значением);

б) нарушенное за счет притока одного нового слова равновесие в лексической системе восстанавливается за счет выхода из системы одного слова.

Выйти может лишь слово из моносемантичной зоны, при этом вероятность выхода у всех слов моносемантичной зоны одинакова и равна $1/n_1$. Переход слова из моносемантичной зоны в зону слов с 0 значениями (т.е. выход из системы) равновероятен переходу слова из моносемантичной зоны в зону слов с 2 значениями, т.е. выход слова из системы в среднем сопровождается переходом одного слова из моносемантичной зоны в зону 2. Вновь нарушенное равновесие восстанавливается за счет выхода (с вероятностью $1/n_2$) двух слов из зоны слов с 2 значениями – одного в моносемантичную зону, другого в зону слов с 3 значениями и т.д.

3. Суммарная картина следующая: слово, попав в лексическую систему, блуждает по законам полунепрерывного случайного блуждания, переходя на каждом такте с уровня на уровень (с более низкого на более высокий либо наоборот) с вероятностью $1/n_m$, либо остается в своей полисемичной зоне с вероятностью $\left(1 - \dfrac{2}{n_m}\right)$. Длительность одного такта в системе задается скоростью поступления новых слов в систему. Верхние дискретные уровни характеризуются малыми значениями $n_m$ и соответственно повышенной вероятностью выхода слов из зоны. Данная черта модели отражает легкость метафоризации диффузных полисемичных значений слов. На самом верхнем уровне полисемии, разрешенном в системе, слово "отражается" назад. Попав на нулевой уровень, слово поглощается и более в систему не возвращается. На данном этапе рассмотрения вопроса количество слов в нулевой зоне полагается бесконечным. При задании конечного количества слов в нулевой зоне появляется возможность численного моделирования феномена возвращения слова в актив, т.е. возрождения в употреблении после периода забвения.

4. Описанная модель позволяет определить основные характеристики жизненного цикла слова – среднюю длительность жизни слова, распределение длительностей жизни слова, распределение слов по степени полисемии в зависимости от их возраста, а следовательно, и зависимость средней полисемии от возраста, распределение возраста слова в зависимости от количества его значений, а следовательно, и вероятный возраст слова в функции количества его значений и др.

5. Была исследована лексическая система русского языка объемом 93 000 слова с распределением слов по полисемичным зонам согласно модели из [1, 2]. Суммарное количество значений всех слов составляет 138 000. Данный язык примерно соответствует модели русского языка, представленной толковым словарем русского языка под ред. Л.П. Евгеньевой и по объему соответствует экспериментально определенному А.А. Поликарповым и А.О. Поликарповой словарному запасу коренного носителя русского языка с высшим (неполным высшим) образованием. На рис. 1 приведена плотность распределения длительности жизни слова (жирная линия). По оси абсцисс отложено время в характерных единицах времени системы – средней длительности жизни слова. Исходя из условия стационарности данная длительность, измеряемая в тактах, численно равна количеству слов в системе. Данная зависимость не может быть аппроксимирована экспонентой, поскольку показательная зависимость срока службы характерна для систем без "памяти". Грубо зависимость может быть аппроксимирована суммой двух экспонент, обозначенных на рис. 1 тонкими линиями.

6. Средняя степень полисемии слова растет с его возрастом и асимптотически стремится к 1,80, что больше средней полисемии в системе – 1,48. Распределение слов определенного возраста по количеству значений отличается от распределения слов в системе в целом. Однако существует предельное распределение подобного рода, к которому стремятся слова по мере их старения, т.е. подсистема старых слов характеризуется отсутствием памяти.

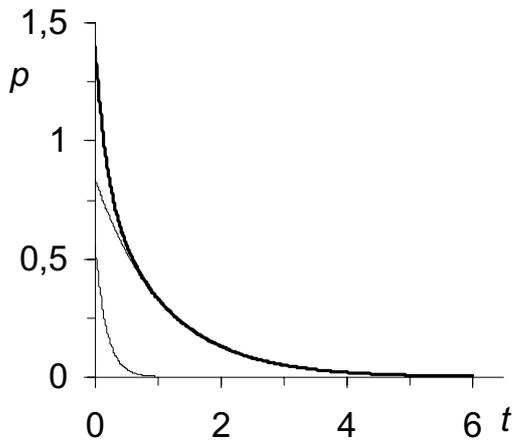 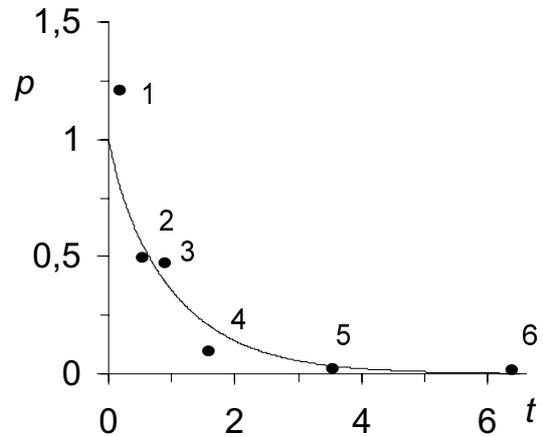

Рис. 1.                                   Рис. 2.

Существует устойчивое предельное распределение возраста слов в системе, не зависящее от начального распределения возраста и определяющееся распределением длительности жизни слова. Вероятностные методы теории восстановления позволяют найти предельное распределение возраста на основе распределения длительности. На рис. 2 непрерывной линией представлено распределение возраста слов в системе. Экспериментальная проверка модели осуществлена по данным о распределении слов русского языка по различным возрастным периодам (табл. 1) [3, с. 23]. В таблице ниже приведено эмпирическое распределение слов по возрасту.

| Период | 20 в. | 19 в. | 18 в. | 15-17 вв. | 7-14 вв. | общеслав. | индоевроп. |
|---|---|---|---|---|---|---|---|
| Средний возр., лет | 50 | 150 | 250 | 450 | 1000 | ~ 1800 | ? |
| К-во слов | 5979 | 14375 | 13758 | 8361 | 5868 | 3619 | 174 |
| К-во слов/век | 35000 | 14375 | 13758 | 2787 | 733 | ~ 450 | ? |
| № точки на рис. 2 | 1 | 2 | 3 | 4 | 5 | 6 | – |

По сравнению с таблицей-источником из [3] добавлены вычисленные по середине соответствующего периода средние значения возраста слов и количество слов, сохранившихся в языке до настоящего времени с разбивкой по векам. Анализ данных таблицы позволяет сделать предположение, что 20 век еще не "закрыт" лексикографически и истинное количество слов, вошедших в обследуемую лексическую систему в 20 веке и сохранившихся до настоящего времени, гораздо больше. Приняв это количество равным 35 000, вычисляем средний возраст слова в языке – 300 лет, и общее количество слов – около 81 000. Близость этого значения к 93 000 позволяет нанести на рис. 2 точки, соответствующие эмпирическим данным. Разность в объемах модельной и эмпирической лексических систем, а также разная размерность оси ординат для двух групп данных учтена путем нормирования. Диаграмма рис. 2 рассчитана без учета разницы между лингвистическим и историческим временем. При учете неравномерной скорости обновления словаря (согласно М.В. Арапову, с 18 века темпы обновления словаря возросли примерно в три раза) согласие теоретических данных с эмпирическими улучшается для 20 века и ухудшается для периода 7–14 веков.

8. На рис. 3 приведены примеры типичных реализаций поведения слова в системе – профили изменения количества значений у слова, просуществовавшего в системе 1, 2 и 3 характерных единицы времени (соответственно диаграммы а, б и в).

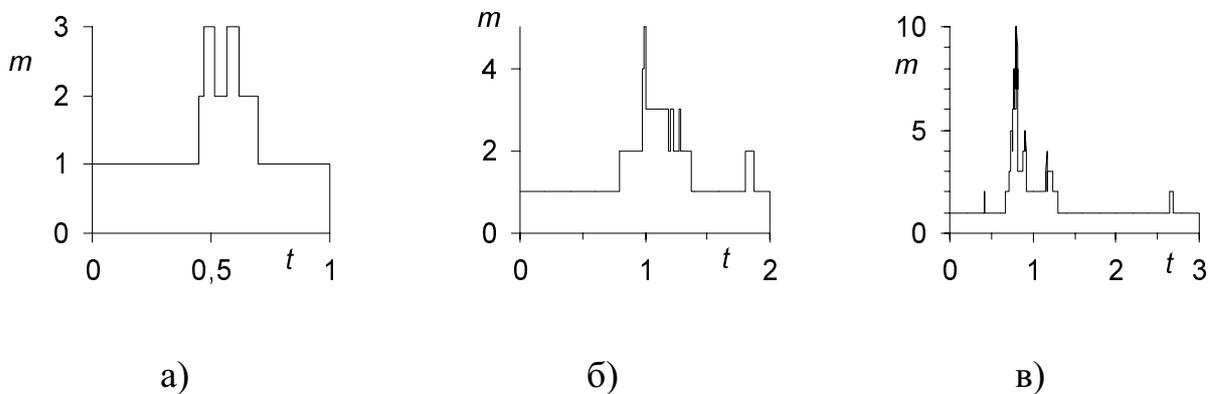

а)            б)            в)

Рис. 3.

Процессы, представленные на рис. 3, смоделированы исходя из принципов случайного блуждания. Общая языковая картина полисемии складывается из единовременных синхронных срезов десятков тысяч подобных процессов.

## Литература